\documentclass[twocolumn]{article}
\usepackage[utf8]{inputenc}
\usepackage[dvipsnames]{xcolor}
\usepackage{multicol}
\usepackage{geometry}
\usepackage{nopageno}
\usepackage{fancyhdr}
\usepackage{amssymb}
\usepackage{amsmath}
\usepackage{wrapfig}
\usepackage{lipsum}
\usepackage{graphicx}
\usepackage{authblk}
\usepackage[skip=2pt,font=scriptsize]{caption}
\usepackage[colorlinks,bookmarksopen,bookmarksnumbered,citecolor=black,urlcolor=blue]{hyperref}

\providecommand{\keywords}[1]
{
  \small	
  \textbf{\textit{Keywords---}} #1
}

\fancypagestyle{plain}{%
\fancyhead{}
 
  \fancyfoot[C]{\thepage}
\fancyfoot[L]{\textbf{\ }}
    \fancyfoot[R]{ \ }
}

\title{Enhancing Human-Machine Teaming for Medical Prognosis Through Neural Ordinary Differential Equations (NODEs)}

\newcommand\BibTeX{{\rmfamily B\kern-.05em \textsc{i\kern-.025em b}\kern-.08em
T\kern-.1667em\lower.7ex\hbox{E}\kern-.125emX}}

\usepackage[backend=bibtex,style=numeric, sorting=none]{biblatex} 
\begin{document}
\makeatletter
\let\@fnsymbol\@arabic
\makeatother
\author{D. Fompeyrine \thanks{\small Founder and CEO, PhD in Clinical Psychology, Myndblue}}
\author{E. S. Vorm \thanks{PhD in Human-Computer Interaction, Lead Evaluator, DARPA Explainable AI Program}}

\author{N. Ricka\thanks{Lead Data Scientist, PhD in Mathematics, Myndblue}}
\author{F. Rose\thanks{Biomedical Data Scientist, PhD in Computational Biology, Myndblue}} 
\author{G. Pellegrin \thanks{Machine Learning and Project Engineer, MSc in Applied Physics, Myndblue}}

\affil[]{ \ }

\date{February 2021}

\maketitle

\begin{abstract}
\textbf{Machine Learning (ML) has recently been demonstrated to rival expert-level human accuracy in prediction and detection tasks in a variety of domains, including medicine. Despite these impressive findings, however, a key barrier to the full realization of ML's potential in medical prognoses is technology acceptance. Recent efforts to produce explainable AI (XAI) have made progress in improving the interpretability of some ML models, but these efforts suffer from limitations intrinsic to their design: they work best at identifying why a system fails, but do poorly at explaining when and why a model's prediction is correct. We posit that the acceptability of ML predictions in expert domains is limited by two key factors: the machine's horizon of prediction that extends beyond human capability, and the inability for machine predictions to incorporate human intuition into their models. We propose the use of a novel ML architecture, Neural Ordinary Differential Equations (NODEs) to enhance human understanding and encourage acceptability. Our approach prioritizes human cognitive intuition at the center of the algorithm design, and offers a distribution of predictions rather than single outputs. We explain how this approach may significantly improve human-machine collaboration in prediction tasks in expert domains such as medical prognoses. We propose a model and demonstrate, by expanding a concrete example from the literature, how our model advances the vision of future hybrid Human-AI systems.} \\
\end{abstract}

\vspace{.25cm}
 \\ \vspace{0.25cm} \\
{\small Corresponding address: \texttt{publishing@myndblue.ai}}

\section*{Introduction}

Businesses and governments around the world are racing at breakneck speeds to build systems that can leverage machine learning (ML) to gain strategic advantages. While ML is quickly being introduced in new fields such as logistics \cite{logistics} and agriculture \cite{ag}, one domain is already an old familiar friend: healthcare. The vision of being able to accurately predict a patient's medical trajectory by integrating vast amounts of disparate data has inspired generations of computer scientists and has resulted in a variety of early applications of artificial intelligence in the form of decision aids and decision support systems. At the heart of this vision is a collaboration between human and machine; a synergistic hybrid system that affords humans with near superhuman abilities to compute massive amounts of data and with it project far into the future with brilliant accuracy. This vision of human-machine teaming through the application of human-AI agents is fast becoming a reality today, thanks largely to ML. Indeed, ML algorithms have proven to be as accurate or better than expert-level predictions in various medical domains, from image classification to time-series analysis, and many others \cite{cancer, pallative}. But while these advances promise much, the realization of true human-machine teaming  in medical prognoses may be hindered by a familiar and stubborn barrier— the lack of human trust. As early as the 1980s, thorough comparisons between computer-generated recommendations and experts had already demonstrated the critical usefulness of artificial decision aids \cite{MYCIN}, but the lack of algorithmic transparency caused significant conflicts and inevitable delays that ultimately prevented the widespread adoption of expert systems into mainstream use. A close reading of the literature from this era reveals that these failures were caused by flaws in \textit{usability}, not by algorithmic accuracy or efficiency. 

Modern day ML algorithms are direct descendants of these expert systems of the past, and they carry many of the same challenges. Concerns over low algorithmic transparency and the blackbox nature of algorithms such as deep learning have given rise to new interdisciplinary fields of research aimed at improving interpretability and transparency of ML algorithms, so-called explainable artificial intelligence (XAI) \cite{Arrieta}. Perhaps driven by lessons learned from earlier generations of clinical decision support failures, XAI has quickly been offered up as \textit{the} solution, even when the problem is seldom articulated or perhaps not even fully understood. In order to better understand why explainability and transparency play a central role in the potential widespread adoption of ML, the following section we illustrate two general scenarios that motivate their importance. Following this, we discuss why XAI alone is insufficient in achieving the goal of human-AI cooperation for medical prognoses. We then introduce Neural Ordinary Differential Equations (NODEs) as a proposed machine learning architecture for use in medical predictions and prognoses, and we illustrate how their use is intrinsically designed for maximum usability, and is superior in supporting human intuition and decision making in medical prognoses. 
\subsection*{The Utility of Explainability}
Research has uncovered two predominant situations in which users of machine learning encounter usability conflicts and hence hesitate to trust their outputs \cite{vorm}. The first are conflicts that arise when an ML algorithm or the overarching intelligent systems that embody them suddenly behave unpredictably or erratically. These off-nominal behaviors can have widespread consequences on user confidence and trust. When systems that are ordinarily predictable and reliable suddenly behave unpredictably or give an unexpected output, questions and concerns naturally follow. Machine learning models that perform very well under one condition often display wildly different behaviors when even small changes are made. Sometimes these errors can be traced to a root cause and behaviors can be easily explained. In other cases, tracing the error is much more difficult, and oftentimes impossible. This is of immediate concern for makers of industrial-scale autonomous systems such as self-driving cars \cite{cars}, but also of great concern in applications that feature machine learning in the role of decision support, as is the case in clinical decision support systems. Physicians, hospital administrators and even government regulators, upon seeing the apparently brittleness of ML are likely to ask themselves "if this system has such low reliability and unpredictability, how can I ethically justify using it for my patients?" Without some measure of assurance of its reliability, low trust and in some cases abandonment of the technology as viable remains the most likely outcome. The argument for XAI is therefore driven by the understanding that \textit {no trust = no use}. Hence much work has recently focused on improving the transparency of ML algorithms to understand \textit{why they fail}. While XAI research has resulted in a number of small breakthroughs in terms of ML development techniques, the true benefits from these efforts are limited mostly to programmers and debuggers whose goal is to build more robust and reliable systems. While important, XAI's current focus on explaining \textit{what went wrong} does little to help users determine when and why an algorithmic prediction may be correct, and so does little to help users determine whether or not to use, trust, engage with, or adopt AI moving forward. To have a measurable effect in these areas, we need a \textit{prospective} focus.

The second motivating scenario for XAI, therefore, is one that arises in situations where the user must make a \textit{prospective} decision based on the output of the system. For instance, this might come in the form of whether or not a radiologist decides to accept or validate a diagnostic flag created by ML on a medical image, or whether or not to act on ML-based predictions that indicate an aggressive treatment regimen may be warranted in a given patient. In these situations, users are not afforded the luxury of ground truth, i.e., there is no direct way of knowing whether or not the ML algorithm is accurate because it is projecting a future state that has yet to occur. Instead, users must wrestle with whether or not a projection of future events (e.g., prognosis) seems likely and plausible. As with any decision scenario of any importance, humans naturally seek additional information with which to inform and support their decision. This information-seeking behavior typically comes in the form of questioning \cite{foraging}, such as \textit{what specific data points predict this person will make a good recovery?} or perhaps \textit{what is the reasoning behind this suggestion to treat with an experimental drug?} The argument for XAI to address this prospective scenario, therefore, is that the more answers to user questions a system can provide, the greater the degree of trustworthiness the system has, and the greater the likelihood that the system will be used to the extent and in the manner in which it was designed. 

\subsection*{Limits to Prospective XAI}
Unfortunately, while the majority of XAI has focused on post-hoc explanation strategies, even the few efforts that are prospective in nature are severely limited in their ability to improve usability and technology acceptance for ML for at least three reasons. Firstly, there are practical limits to how many questions a system can answer, or how much information can be meaningfully provided to human users. Designing systems that seek to provide mappings to every component and sub-component would be cumbersome to the point of being unusable. While rules governing a natural phenomenon's evolution lie in a constrained space that can hypothetically be modelled completely, a fully transparent XAI prediction would need to be able to master all possible future states even in regions that do not seem plausible at first. This seems wildly unrealistic, as it would require a dataset of unattainable size to explore and understand all the possible configurations. Labyrinthine causal diagrams of high dimensional datasets are simply impractical as they are too difficult (or impossible) for a human to interpret. 

Secondly, another limitation to XAI approaches stems from how humans reason about causality. Cognitive scientists have long demonstrated that humans do not typically engage in the kind of deliberate, methodical decision making (i.e., “slow thinking,” or “system two thinking”) that would make use of such a robust and complete XAI system. Rather, most decision-making strategies are predominantly those that make efficient use of heuristics, or mental shortcuts (i.e., “fast thinking” or “system one thinking” \cite{khaneman}). Human cognition strategies have evolved largely to prioritize rapid decision making. Most decision making scenarios are those where humans make quick assessments of the information and act, rather than cautiously and systematically pour over all available data. In other words, more data is seldom likely to result in better decisions.

Lastly, a limitation of XAI in improving the prospective prediction problem is that developers maximize predictive accuracy of ML models, but do little to address the myriad of other human factors that play a role in how humans prognose and make decisions. The role of intuition in expert decision making has received much focus in the cognitive and neurosciences for many decades, especially in tasks such as discovery and exploration \cite{int, int2, int3, int4, int5}. Earlier generations of artificial decision aids that attempted to mimic human decision making ran into trouble because they could not account for information originating from outside of their knowledge base. Developing expert knowledge seems an illusive target for an artificial system because, as human expertise grows, it also evolves towards more and more intuition \cite{Pa18} and subjectivity, and draws conclusions from information that is broader than merely the data in a patient's medical record. How patients look, how they speak, and how family members interact with them are all examples of factors that could potentially inform an expert clinician and contribute to their decision strategy. This extra-cognitive information is both difficult to characterize and difficult to model in ML. Current ML strategies do not prioritize or make use of human intuition in their predictions, and so explanation strategies are not likely to improve the likelihood of experts using them. Any cooperative vision where ML is a trusted component in a cooperative decision making system, such as a fully-integrated clinical decision support system, should feature the strengths of both components (human and machine), rather than limiting the strengths of one over the other.

Although explainability is a vital factor in affecting human trust in ML algorithms \cite{trust}, it is not entirely sufficient to achieve the true vision of how ML can help humanity by improving our ability to predict the future. To achieve true human-machine collaboration, especially in expert domains where high levels of risks are inherent, ML systems will need to do more than merely explain themselves. They will also need to adapt to and in some cases overcome the natural limitations of our human cognitive evolution. 

\subsection*{Horizons of Predictability: Limits of human cognition in prediction} 
\begin{figure}[h!]
    \centering
    \includegraphics[width=0.5\textwidth]{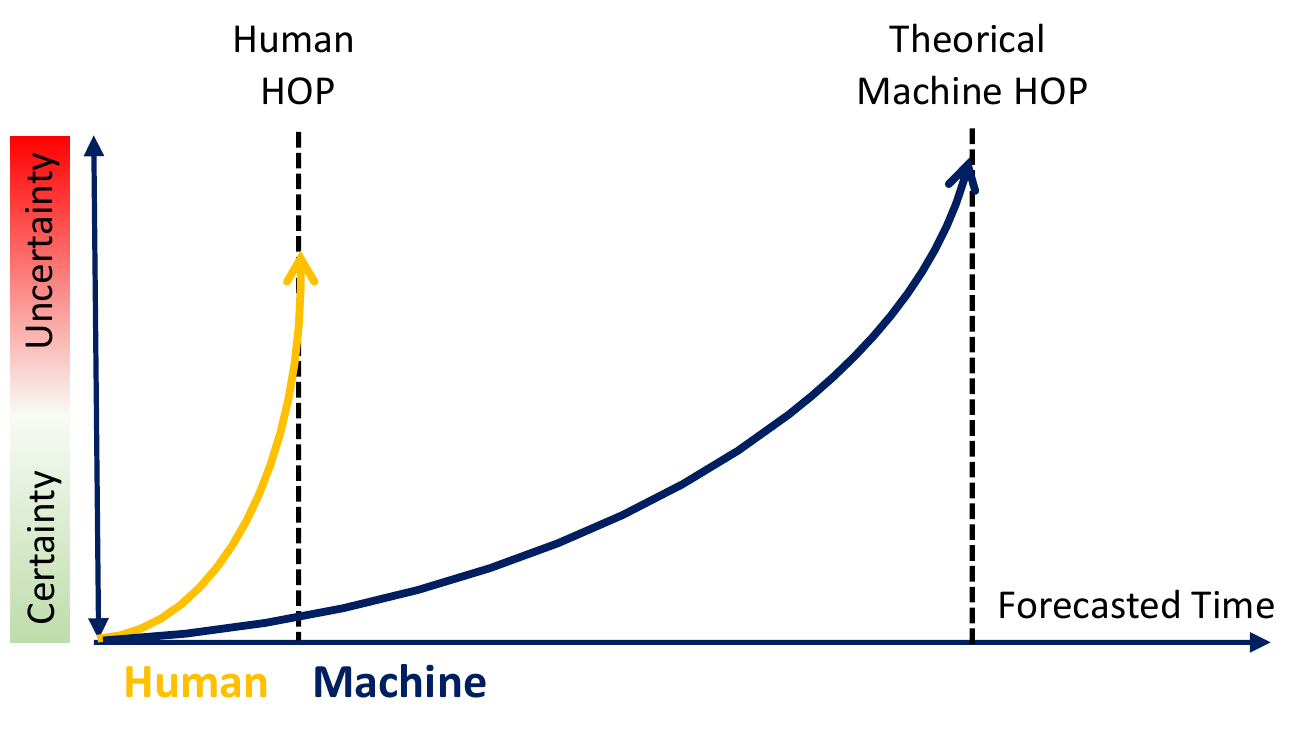}
    \caption{Human and machine horizons of predictability (HOP). Machine learning is able to make an accurate prediction at a longer timescale than human beings, but humans often struggle to trust ML outputs because they are difficult to comprehend, and do not incorporate all available information, including human intution. Our proposed architecture extends human predictive performances up to a time nearer to the theoretical machine HOP, thus enhancing human-machine teaming in medical prognoses.}
    \label{fig:HOP}
\end{figure}

In the field of physics, the \textit{horizon of predictability} (HOP) refers to the limit after which forecast becomes impossible due to the exponential accumulation of errors \cite{Str19, solar20}. Machine learning has a similar limit to its predictive horizon for the same reason  \cite{solar89, solar90, solar92, solar20}. This limit is unbreakable, in the sense that even with perfect knowledge of the underlying dynamics of the system, it is impossible to make predictions beyond a certain point because the latent errors compound to such an extent that no certainty can be achieved. Although there are limits to how far out ML can accurately make predictions, that horizon of predictability extends far beyond the horizon of predictability of human beings (e.g., Figure~\ref{fig:HOP}). 

\begin{table*}[h!]

\includegraphics[width=\textwidth]{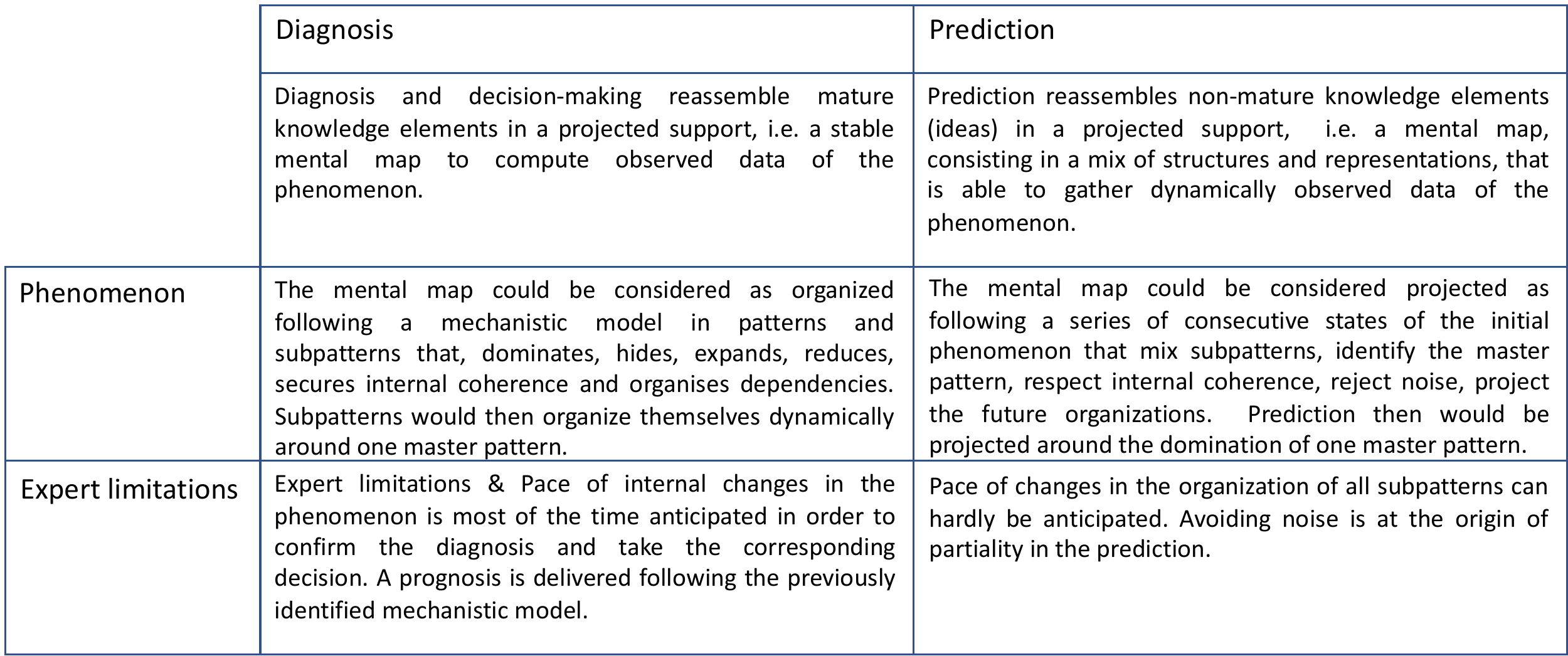}
\caption{Structure of mental process for diagnosis and prediction of a human expert.}
\label{tab:limitations}
\end{table*}
The limits of human prediction stem mainly from our own cognitive capacity and tendencies, rather than from latent errors in the data. Limits to human cognitive capacity are well known. For example, Miller’s Law, or the so-called ‘magic number 7 plus or minus 2’ illustrates the limits of working memory functions of human beings \cite{miller}. Humans have other well-known computational challenges as well. For example, they often struggle to comprehend abstract concepts such as single-event probabilities and non-linear distributions of data \cite{logs}. These cognitive limitations severely limit human ability to make accurate predictions, creating in essence a very near horizon of predictability. 

Aside from these computational limitations, humans also suffer from cognitive flaws that limit our ability to accurately project future states. As mentioned earlier, our understanding of human evolution points to the prioritization of rapid pattern recognition, but not necessarily the ability to uncover and explore new emerging patterns. Our instinct to focus on single dominant patterns is quite useful in identifying and classifying known entities (e.g., diagnosing). But this instinct also means that our ability to predict future events is ultimately fragile because our focus on identifying dominant patters often means that we exclude emerging sub-patterns (what is necessary to accurately make a prognosis). The process of \emph{Diagnosing} \cite{HKA11} requires a mechanistic model, which necessitates multiple knowledge fundamentals at different levels of maturation. This information is used to guide our exploration until we find an eventual matching pattern, and hence a diagnosis is confirmed. The primary mechanism through which diagnoses are made, however, is through a "ruling out" process, which consists largely of seeking evidence to support a main hypothesis, and systematically dismissing other hypotheses that are not supported by the data. 

\emph{Prognosis}, on the other hand, requires us to admit the projection of ideas not yet formalized on a representational support, i.e. a mental map that has not matured to a full mechanistic model. In an attempt to separate informational uncertainty from intrinsic medical uncertainty, experts naturally attempt to anticipate future changes. Unfortunately, this projection suffers from the same confirmatory bias as mentioned before \cite{WLL15}. When attempting to make predictions, research demonstrates that the projection of a series of consecutive states of a phenomenon is usually ruled by a dominant master pattern, to the exclusion of other potentially informative and influential patterns \cite{Pa18}. This dominant pattern is heavily informed by a feeling of coherence, which is affectingly charged before been conscientiously represented \cite{Luu10}. In other words, to make sense of the chaos, human beings tend to arrange available information into a form of a narrative \cite{narrative}. Studies consistently show that decision making is greatly influenced by how coherent a person's narrative is constructed— whether that narrative is self-chosen, or presented to them in the form of “evidence” \cite{laws}. To determine a prognosis, therefore, the prognosis that seems most likely and plausible to the person is the one that arranges the data in the most coherent structure— i.e., the one that tells the most convincing story. Unfortunately, as has been demonstrated before, data do not always arrange themselves neatly into logical causal relationships that can be quickly appreciated by human beings, which sadly means that a great deal of the time, human beings have a tendency to see connections where there are none \cite{explanation}. In summary, our evolutionary drive to seek dominant patterns and our affinity to arrange data into a narrative format is especially useful when it comes to diagnosing, but not especially useful for making prognoses.
In order to achieve true human-machine collaboration where experts confidently leverage the predictive power of ML, the task at hand, therefore, should not be to focus solely on creating more predictive algorithms, or creating more explainable models. These efforts have already demonstrated their futility through previous generations of clinical decision support systems. What we need instead is to create human-machine systems that allow for the uniqueness of expert human intuition to combine with the distant horizon of predictability of machine learning.

\section*{A Post-Explanation Paradigm Shift}

So far we have detailed the problems that may create usability conflicts between users and machine learning algorithms. Despite highly accurate systems, these conflicts pose a significant threat to the likelihood of machine learning integrating and being formally adopted by expert domains such as medicine. Because machine learning can reason and project out much further than human capabilities, there is a gap between the machine and human horizon of predictability— the limits at which accurate predictions can be made. Current XAI approaches alone will not narrow this gap because a) they are mostly retrospective in focus and do very little to explain future predictions; and b) we have human cognitive limitations (i.e., we have a tendency to focus on predominant patterns that are familiar to us and therefore ignore emerging new patterns, and we have cognitive limitations in how much data we can process). 

To overcome these limitations, we need systems that are specifically designed with the human predisposition for cognitive intuition in mind in order to enhance acceptability and encourage collaboration. A system that seeks to augment, as opposed to supplant, intuition would be one that presents its outputs in forms that are \textit{easily understandable}, to the point of being practically available for humans to use as part of their reasoning. We cannot expect all users of ML to become experts in computer science in order to use ML. Nor do we not want AI that presents itself as an oracle, or one that requires humans to trust it implicitly and not ask many questions. But we also must be mindful of not creating “coercive AI” or “persuasive AI” that lead human decision makers down a path of our own choosing. So what are we to do? 

Rather than developing ways to extract information from intractable models, a plausible solution to encourage better human-machine collaboration with ML is to design machine learning in such a way that its mathematical forms and representations maximize human understanding and comprehension. Rather than requiring humans to understand the mechanisms underlying ML, why not develop ML in such a way that its outputs are packaged in a format that most humans can naturally understand? Much research has demonstrated that the way information is \textit{represented} (i.e., how it is displayed and visualized) can determine a great deal on whether or not humans will comprehend and understand it. For instance, the statement "If a patient has COVID-19 the probability that they will have a positive result on a rapid test is 95\%" is often confused with "if a patient has a positive test result the probability that they have COVID-19 is 95\%." This is an example of how causality, the direction of inference, and conditional probabilities can easily be confounded. In the example above, the first statement is referring to the sensitivity of rapid COVID-19 tests (95\% accurate at detecting COVID-19 \cite{covid}). The second statement, however, confounds the directional inference, mistakenly reversing the conditional probability \cite{frequencies}. For this reason, best practices when displaying statistical risk call for the use of frequency statements (e.g., COVID-19 tests will successfully identify 9 out of 10 people who are infected), as they are more intuitively understood by most people \cite{risk}. 

Another example, one salient to ML, is the reliance on probabilities to communicate uncertainty. This strategy is very problematic for a variety of reasons. First, humans do not understand probability very well unless they are specifically trained to do so \cite{risk}. Second, in order to fully appreciate probability, it is necessary to have information related to base rate and frequency of occurrences (something that is seldom afforded to users). Thirdly, single-event probabilities are notoriously prone to being misunderstood by users \cite{logs}. For example, the statement "The system is 40\% certain that a patient will develop PTSD" can be interpreted a number of different ways. One might interpret the statement to mean that 40\% of patients with profiles like this one will develop PTSD, while another might interpret the statement to mean that the system will be able to predict future PTSD in 40\% of patient records. These are all simple examples of how the way that information is represented, or its \textit{form}, can either make that information better understood, or more likely to be confused. Just as numbers can be expressed in a variety of different forms, the outputs of ML can also. Our approach to using mathematical representations that capitalize on and augment human intuition is to use Neural Ordinary Differential Equations (NODEs) \cite{CRBD18}.

\subsection*{Neural Ordinary Differential Equations: An elegant solution to the paradox of explainability}

Ordinary Differential Equations (ODE’s) are well known in the fields of applied and pure mathematics. Their long history of beneficial use in physics and engineering has resulted in large and extremely well-tested, high performing differential equation libraries. Differential equations are a tried and tested tool for modelling data that until 2018 had been largely left out of the conversation surrounding machine learning. Their introduction as an architecture for machine learning was met with much surprise and critical acclaim from the scientific and computational communities of practice, including the best paper of the year at the 2018 Conference on Neural Information Processing Systems (NEURIPS, \cite{CRBD18}). 

Applied to machine learning, Neural Ordinary Differential Equations (NODEs) are algorithms that encode the dynamics of a system by learning an ordinary differential equation for function approximation, as opposed to training a neural network. NODEs have several advantages over other machine learning techniques for providing clear and tractable outputs. First, they express the solution in continuous time as opposed to models discretizing the timeline into small time steps \cite{Duv20, Str20, Symoden, KZ20} and can learn on irregular time-series to best match real-world data (for instance biological measurements in the medical field). As opposed to the more common Partial Differential Equations (PDEs) \cite{CRBD18, RCD19, Duv20}, where the dynamics of a multi-variate function is modeled; NODEs only consider differentials with respect to a single parameter \cite{PDEa, PDEb, PDEc}. Because we are interested in future projections (i.e., predictions or prognosis), the most relevant continuous indexing parameter is \textit{time}. Consequently, we posit that using NODES with all derivatives being with respect to the time variable will afford users a tremendous benefit in being able to comprehend and trust ML  outputs for future predictions. For instance, using a NODE architecture, it is possible to let the latent information evolve for an arbitrary long time to uncover subtle information about the future evolution of the system. This serves as a useful method of simulating future states, with time as the single differentiating factor. Similarly, NODEs can be used to invert the arrow of time, and effectively reproduce the steps they took to arrive at any observed state of the system. This effectively affords users a traceability analysis, and allows users to answer questions about the steps that led to the current observed state of the system. This process is described in the first line of Table \ref{tab:shift}.

In addition to the benefits mentioned above, NODEs also show better long-term predictions than classical recurrent neural network (RNN) architectures. Published works \cite{CRBD18, RCD19} (and the companion code \cite{LatentODEGithub}) have demonstrated for the first time the use of NODEs in a latent ODE architecture to model patients' trajectories from physiological data recorded in an intensive care unit (ICU). In this work, NODEs show a better sequence reconstruction and state-of-the-art accuracy when predicting in-hospital mortality or risk of re-admission compared to other deep learning architectures \cite{RCD19, ICUBenchmark}. More broadly, a system based on NODEs could be especially well suited to predict future states in noisy dynamic systems, such as those commonly found in clinical decision support.

We summarize in Table \ref{tab:framework} the main improvements between existing explainability frameworks and our proposed approach using NODEs.

\begin{table}[h!]
    \centering
    \includegraphics[width=0.75\textwidth]{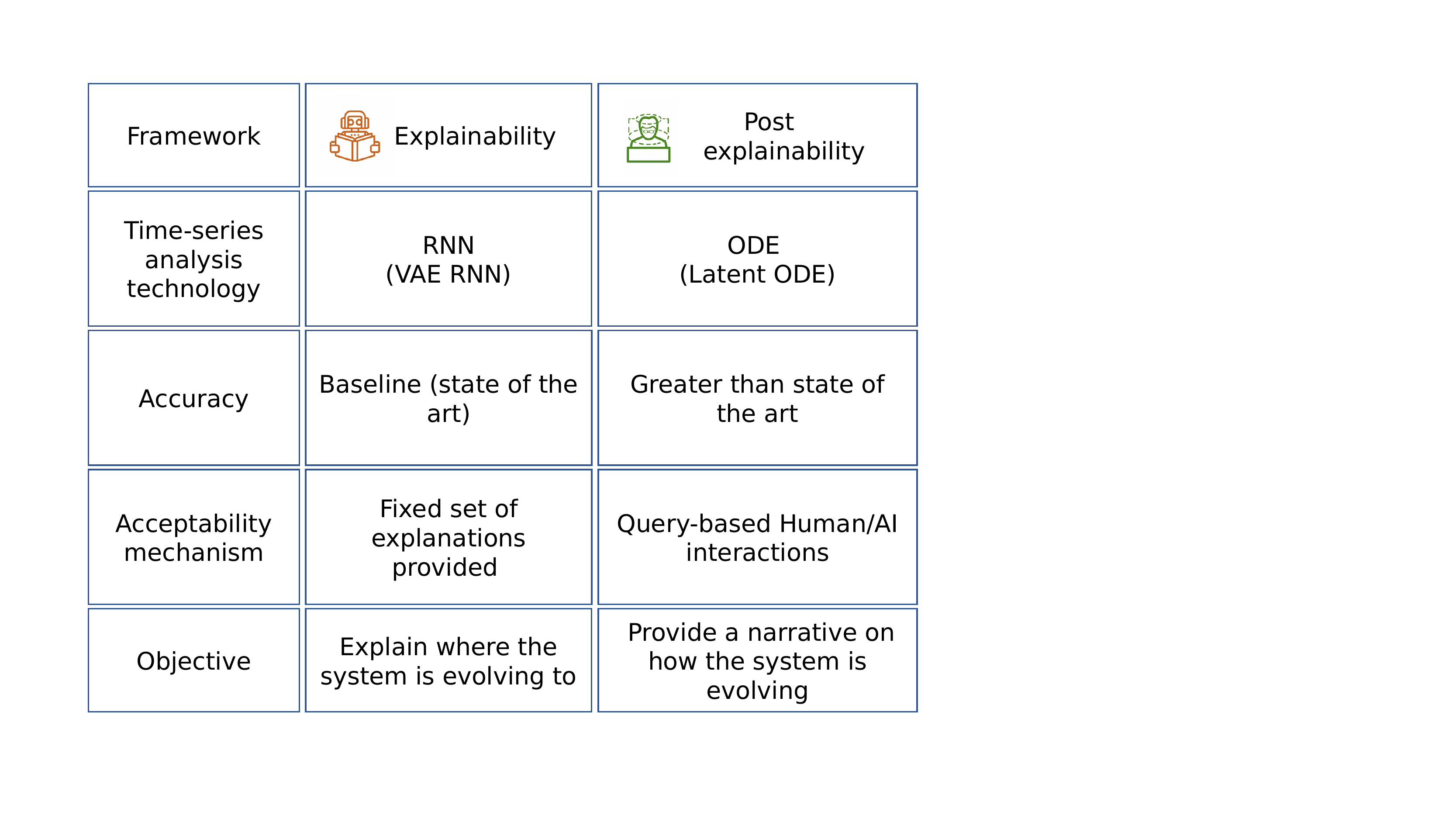}
    \caption{Key changes between the explainability framework and the post-explainability framework presented here.}
    \label{tab:framework}
\end{table}

\subsection*{Properties of the latent space modelled by latent ODEs}
To briefly illustrate and summarize the basic function of NODEs, we will briefly discuss latent ODEs and their technical structure. Latent ODEs are used to model the evolution of a process across a time series based on data from an initial latent state. While RNNs are the go-to solution for modeling regularly sampled time-series data, they do poorly when presented with irregular or inconsistent data, such as the data commonly found in a patient's medical record. To achieve success with traditional RNNs when dealing with inconsitent or irregular time-series data, many workaround steps in data preprocessing are necessary \cite{CRBD18}. These steps result in fairly accurate predictions, but without any of the information (particularly the time-related information) necessary to understand the latent variables underlying the prediction. Latent ODEs, on the other hand, are superior to traditional RNNs because they are flexible with respect to incomplete or inconsistent data, and are especially capable at modeling the future across time. The resulting latent trajectory should contain information that is both useful for the main classification task, and for the reconstruction, thus showing the important features of the original time-series. Accordingly, this architecture is intrinsically suitable for irregularly sampled data, as is common in healthcare data, whereas existing approaches must add timestamps to RNNs in an artificial way.

Roughly speaking, the latent ODE system takes measurements $(x_0, \hdots, x_t)$ as input, and translates them into a latent internal representation $(z_0, \hdots, z_t)$ with internal dynamics following a learned equation $$\frac{dz}{dt} = f_{\theta}(z, \epsilon),$$ where $f_{\theta}$ is expressed by a deep neural network taking into account the noise $\epsilon$ involved in the system. The whole latent trajectory depends only on $z_0$, and can be extrapolated for an arbitrary long time by integrating the differential equation, giving extrapolations $(z_0, \hdots, z_N)$ for any $N$. Finally, the latent trajectory is decoded into an approximation $(\hat{x}_0, \hdots, \hat{x}_N)$ of the original measurement. The encoder, decoder and differential equation weights are trained so that $\hat{x}$ is as close as possible to the real trajectory $x$. It was previously observed in the literature that latent ODEs achieve results that are comparable or better than state-of-the-art performances on real life datasets (on the MIMIC-II dataset, see table 6 in \cite{RCD19}, reproduced here as Table \ref{tab:rnnVsOde}, and on the MIMIC-III dataset see \cite{ICUBenchmark}). We refer the reader to \cite{CRBD18, RCD19} for more extensive details on latent ODEs in machine learning.

\begin{table}[h!]
\centering
\includegraphics[width=0.45\textwidth]{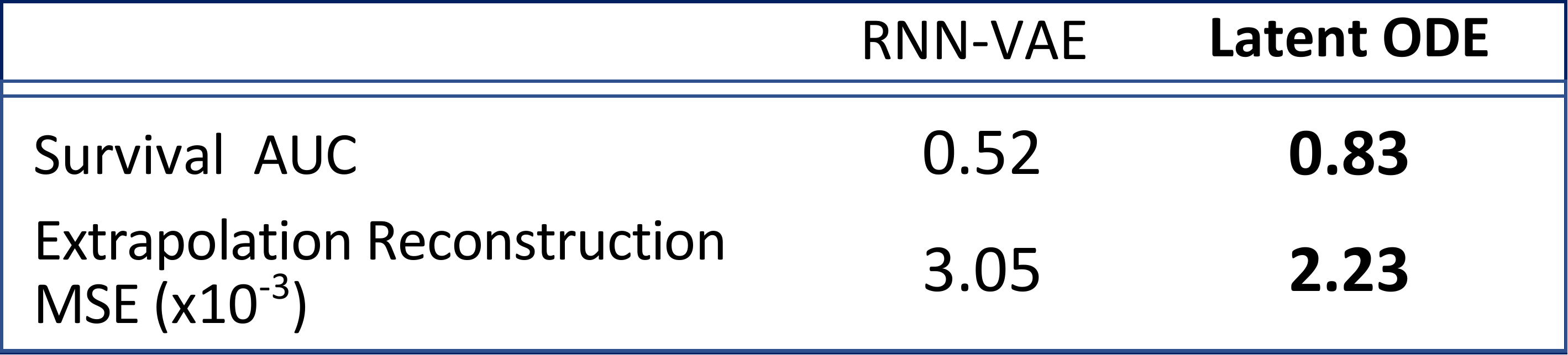}
\vspace{0.2cm}
    \caption{Results of classification and reconstruction for the MIMIC-II ICU dataset. The task is to predict the survival of ICU patients, measured by the survival accuracy and AUC. The goodness of the reconstruction is measured by the mean square error (MSE) on normalized features.}
    \label{tab:rnnVsOde}
\end{table}

Latent trajectories have been demonstrated on simulated datasets in the literature (see the examples on the spiral dataset in \cite{CRBD18}). These analyses, however, need to be interpreted within a certain context. First, simulated examples are usually low dimensional, so generating a visually compelling latent space does not necessary imply that it will be possible for real life scenarios where data is noisy, incomplete, irregularly sampled, etc. Second, the task studied for these simulated examples are usually restricted to reconstruction. Thus it is impossible to question whether the latent trajectory actually supports a prediction. For instance, enforcing acceptability of an automatically generated prognosis by showing the possible futures of the patient and the important changes that will occur during the projected trajectory.

The analysis made in \cite{RCD19} focused on the neural network's ability to predict patient mortality. Our main objective, however, is to show that using NODEs to model a system's evolution leverages \textit{additional information} about a patient's trajectory, which contributes to human-level understandability and therefore improves the acceptability of the output (assuming the output is accurate and deserves to be accepted), while not compromising the predictive power compared to state-of-the-art approaches. 

In the next section, we demonstrate how using a NODE architecture in machine learning can be applied to provide enhanced acceptability and usability. To do this, we demonstrate our approach on a real life medical dataset (MIMIC-II), and analyze to what extent the architecture proposed by \cite{CRBD18, RCD19} helps our purposes. The MIMIC-II dataset is a public dataset with de-identified clinical care data for over 58,000 hospital admission records collected in a single tertiary teaching hospital from 2001 to 2008. In this work, we focus on the mortality task: predicting whether the patient will die in the hospital, and we also produce a study of the reconstruction trajectories from \cite{RCD19} in the case of ICU patients in order to demonstrate how these data dramatically improve the usability of machine learning predictions.

\subsection*{Offering a probabilistic trajectory helps trigger human capabilities}

Due to the probabilistic nature of NODEs, our proposed architecture can afford not only a robust and tractable future patient trajectory, but a \textit{distribution of trajectories,} each representing multiple potential futures of the patient, and each with associated probabilities. (For an illustration, see \cite[Figures 4 and 5]{RCD19}). In practice, this distribution of trajectories would afford the user a great deal of insight. First, the user would be able to easily observe the machine horizon of predictability as the point at which curves are too divergent to extract a coherent behaviour. Traditional RNNs provide no such  indication as to when a prediction becomes untrustworthy, and systems thus must be programmed to rely on training parameters to set a fixed horizon of events independent of the system's dynamics. NODEs, on the other hand, display their horizon of predictability intrinsically and, most importantly, intuitively. Trajectories that lie before this horizon, therefore, are ones the user can have greater confidence in, and each can be analyzed individually. 

It is in the analysis of these potential scenarios where human intuition may be allowed to combine with the predictive power of ML, and in doing so, may flourish. By providing a timeline with a broad array of potential futures, users can explore these potentials in a way that maximizes and prioritizes their expertise AND intuition because they are now afforded access to multiple potential emerging patterns, instead of having a single dominant pattern presented to them. The form that NODEs take, therefore, affords and encourages a kind of "information foraging" \cite{foraging, stopping}, where new emerging patterns are allowed to be considered rather than ruled out preemptively. NODE trajectories also allow for the exploration of various narratives, arranging and displaying data in a format that natively makes sense to human experts. The strengths of NODEs illustrated here- a distribution of trajectories along a timeline that affords easy access to predictive boundaries of the machine while allowing multiple potential future scenarios to be explored- emerge as a natural side effect of the architecture. In other words, in the same way that conveying risk through the use of frequency statements naturally enables people to grasp statistical information and make better decisions, so too do NODE architectures in machine learning. Table \ref{tab:shift} summarizes the main advantages of our hybrid human-AI approach approach with respect to the classical RNN approaches.

To demonstrate these claims, we ran an analysis of the MIMIC-II dataset, which is also studied in \cite{RCD19}. This dataset is quite complex, full of real-life data that is at times noisy, sometimes incomplete, and has much inter-patient variability. These conditions represent many of the characteristics that can harm the predictive power and learning of an ML algorithm, and make interpretation even more difficult. By analyzing this data, we aim to demonstrate to the reader the many inherent strengths of NODE architecture. A discussion of our findings will follow our methodology below. 

\subsection*{A real-life example: ICU patients trajectories}

\begin{figure*}
    \centering
    \includegraphics[width=15.5cm]{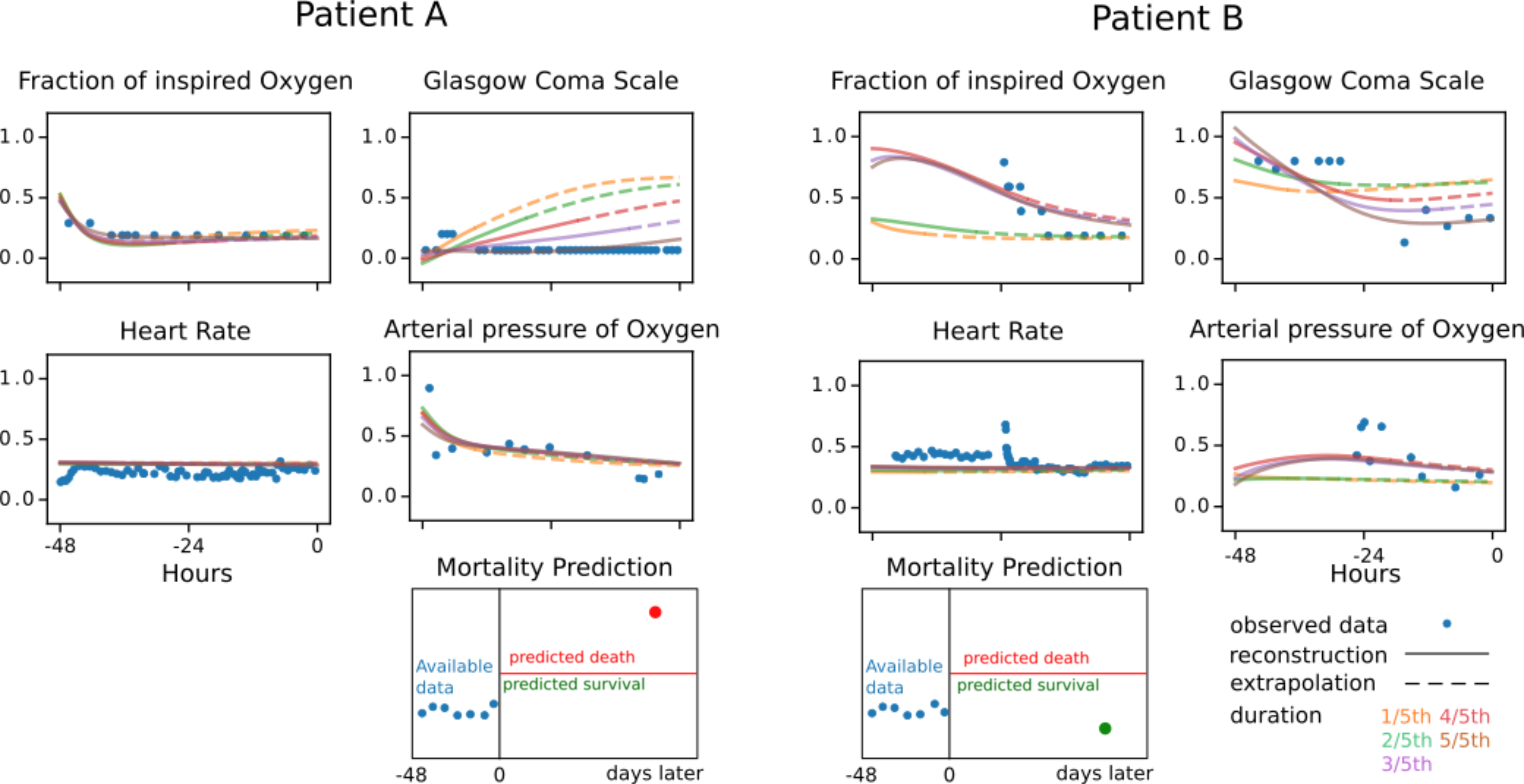}
    \caption{Example results from the latent NODEs model on the ICU data. Medical observation from two patients \textit{A} and \textit{B} are shown in blue for 4 normalized features: fraction of inspired oxygen (FiO2), Glasgow Coma Scale, Heart Rate and Arterial pressure of Oxygen. The different curves show the reconstruction and extrapolation predicted by the model if given a \textit{duration} of 1/5th, 2/5th,... of data from the beginning of the time series. We see that for some features like the Glasgow Coma Scale or the FiO2, reconstructions for patient B tend to follow the tendency of the real feature.
    The \textit{mortality prediction} plot shows the model prediction of the in-hospital mortality. Given the 48 hours of data, the system is able to predict the death of patient A and the survival of patient B several days later.
    }
    \label{fig:results_icu}
\end{figure*}

Our first step was to analyse a slightly modified version of the algorithm trained in \cite{RCD19}. For our study, training time was extended; better and more variable reconstructions were triggered by reducing the noise parameter, thus limiting the power of the encoder power and increasing the internal ODE weights. Two samples (patients) with the two possible outcomes (survival and death) were randomly chosen to study the predictions.

Figure \ref{fig:results_icu} represents a 48 hour window of time. Each box represents a different measurement category (i.e., inspired O2, Heart Rate, etc). The original measurements (blue dots) are displayed. As the reader can see, some measurements are sparser than others. This represents the various inaccuracies and inconsistencies of the data. For example, the arterial blood pressure for patient B is only measured during the second day. Using these measurements, multiple reconstructions, corresponding to the duration of data fed to the algorithm, are conducted for each feature: the solid lines correspond to the reconstructions where original data is known, whereas the dotted lines of the curves correspond to an extrapolated estimation of the patient's future. Multiple dotted colored lines indicate multiple potential futures. 

As we can see, for parts with completely missing observations, the algorithm tends to estimate its values, knowing all the other measured features and the characteristics of the dataset. These curves are not flat, so this does not correspond to an imputation to the mean. Note also that, for these missing features, the algorithm refines the shape of the estimation curve as information grows. Some short-scale variations are not well reconstructed by the latent ODE favoring a smooth curve, as the heart rate peaks around the 24th hour of patient B. This shows a direction to improve current NODE models. 

Take for example, patient A. If we look closely at the Glascow Coma Scale (GCS), we can see that the model initially projects an improvement, as seen by the orange and green curves which correspond to 1/5th and 2/5ths of our 48-hour window (roughly the first 20 hours).  We see, however, that these projections quickly become accurate when enough data is aggregated. The red line projects what might be considered a median outcome, and has a slightly more distant horizon of predictability, while the purple line and finally the brown lines show little or no improvement on the Glasgow Coma Scale. The brown line remains solid throughout the 48 hour window, indicating that high predictive validity and confidence. Because we have overlaid the actual measurements of this patient, we can see that the actual GCS data never improved throughout this 48 hour window, thus validating the brown line's prediction.

The last subplot of Figure~\ref{fig:results_icu} represents the mortality prediction: for each duration of given data a latent trajectory is drawn by the system, from which a simple neural classifier computes mortality chances. For patient A, the mortality prediction stays low at the beginning but rises quickly and ultimately crosses the threshold just before the 48-hour mark, indicating that the system predicts patient A will not survive. We might infer from the data that this prognosis is due to the stable and deteriorated coma state. Although the data shown here is not sufficient alone to make a full cause of death analysis, this simple example demonstrates the ease with which one can access this data and quickly make sense of the underlying connections and their subsequent effects on the predicted outcome. 

\begin{figure*}[h!]
    \centering
    \includegraphics[width=0.95\textwidth]{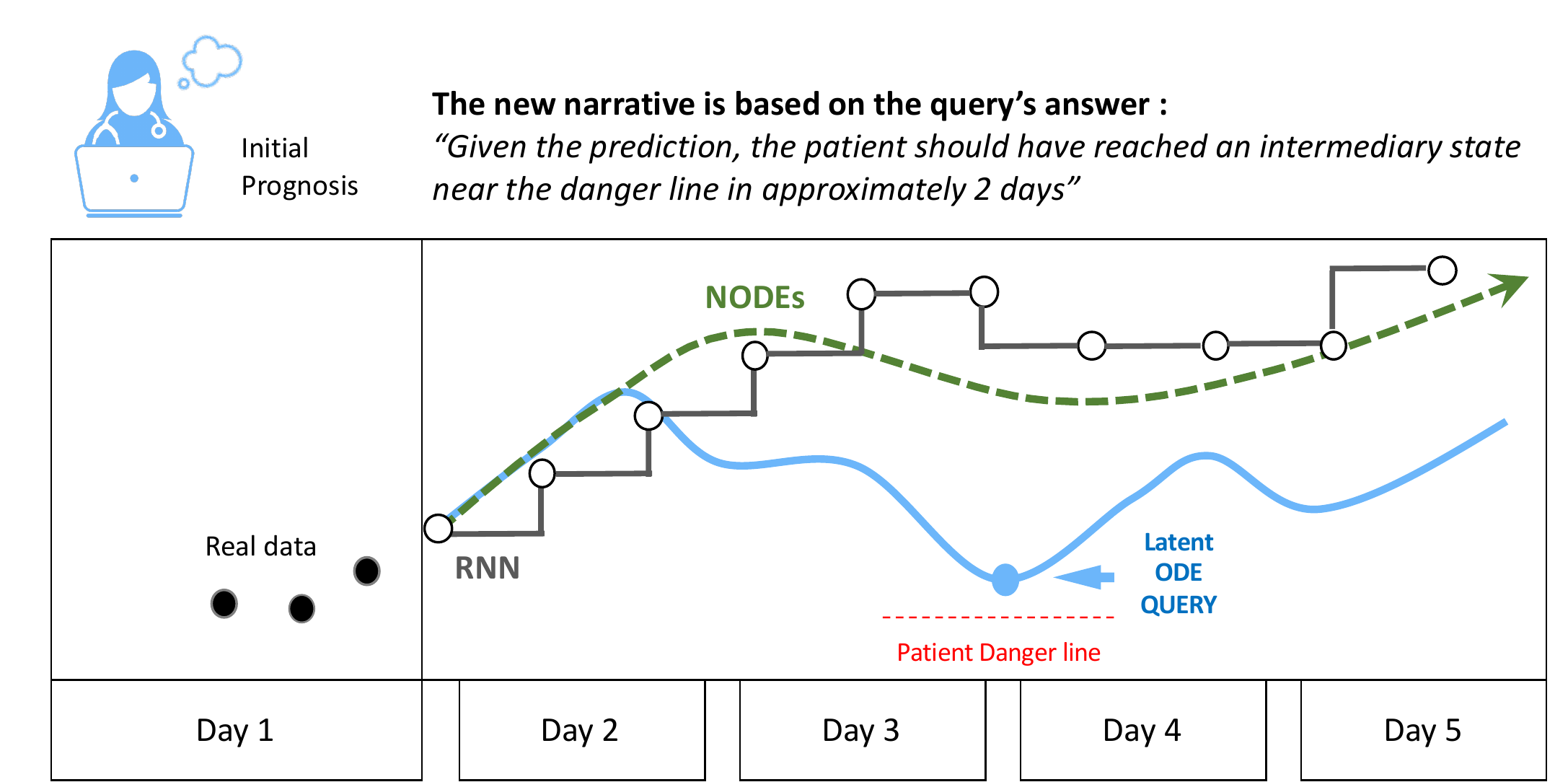}
    \caption{Compared to RNNs, an ODE-based approach produces smooth curves which can be evaluated at any point of the trajectory. Once measured real datapoints are fed to the machine, estimations of its extrapolation can be produced (green curve). The expert user can ask queries such as "what is the trajectory of a patient who gets close to a dangerous situation (blue dot)? The latent ODE then constructs a family of most likely trajectories that passes through this newly added point. This extra-information will help the experts to construct a narrative that is compatible with their knowledge, reinforcing their decision process, or to explore the complex family of possible trajectories by asking more specific queries.}
    \label{fig:draw_probabilistic_advantages}
\end{figure*}

For another example, let us examine patient B. The mortality prediction for patient B remains low and even decreases after 24 hours showing the model's confidence in its prediction. It is important to note that the mortality predictions are being made as new data arrives across this 48 hour period. Along those 48 hours are modelled events (i.e., reconstructions) that originate directly from the ODE architecture. Both the reconstructions and the mortality predictions demonstrated here illustrate that the latent ODE architecture can handle complex sparse real-life data in a manner that is human-understandable and intuitive, while remaining highly accurate.

\section*{Conclusion} \label{sec:discussion}

In the previous section, we illustrated that the NODE architecture is capable of reconstructing a real life dataset, and have demonstrated how an expert might explore the data and produce a narrative in accordance with the NODE's results and predictions. When attempting to make a prognosis, the ability to visualize in detail the system's future evolution aids the expert in generating a narrative about the system. The ease of use afforded by NODEs, combined with multiple future projections provide simple but powerful insights that extend the human horizon of predictability beyond normal limits, and does so in a way that minimizes bias and maximizes trust in the data. 

The latent ODE architecture afford the user the possibility to add new hypothetical measurements in the future, and enable the user to ask the system for the most probable paths that led there. For instance, the expert might choose a specific curve that leads to a region of the feature space that is close to a dangerous situation, and make the following query: "if the system crosses the frontier of the dangerous region, what happens next, and how did the system evolve to end up here?". This is depicted as the blue line in Figure \ref{fig:draw_probabilistic_advantages}. As you can see in this figure, the system that ends up close to the dangerous region at the end of the third day does not cross the frontier with this region, so the user may be confident that this situation is not a concern.

\subsection*{Towards augmented decision-making}

\begin{table*}[h]
    \centering
    \includegraphics[width=\textwidth]{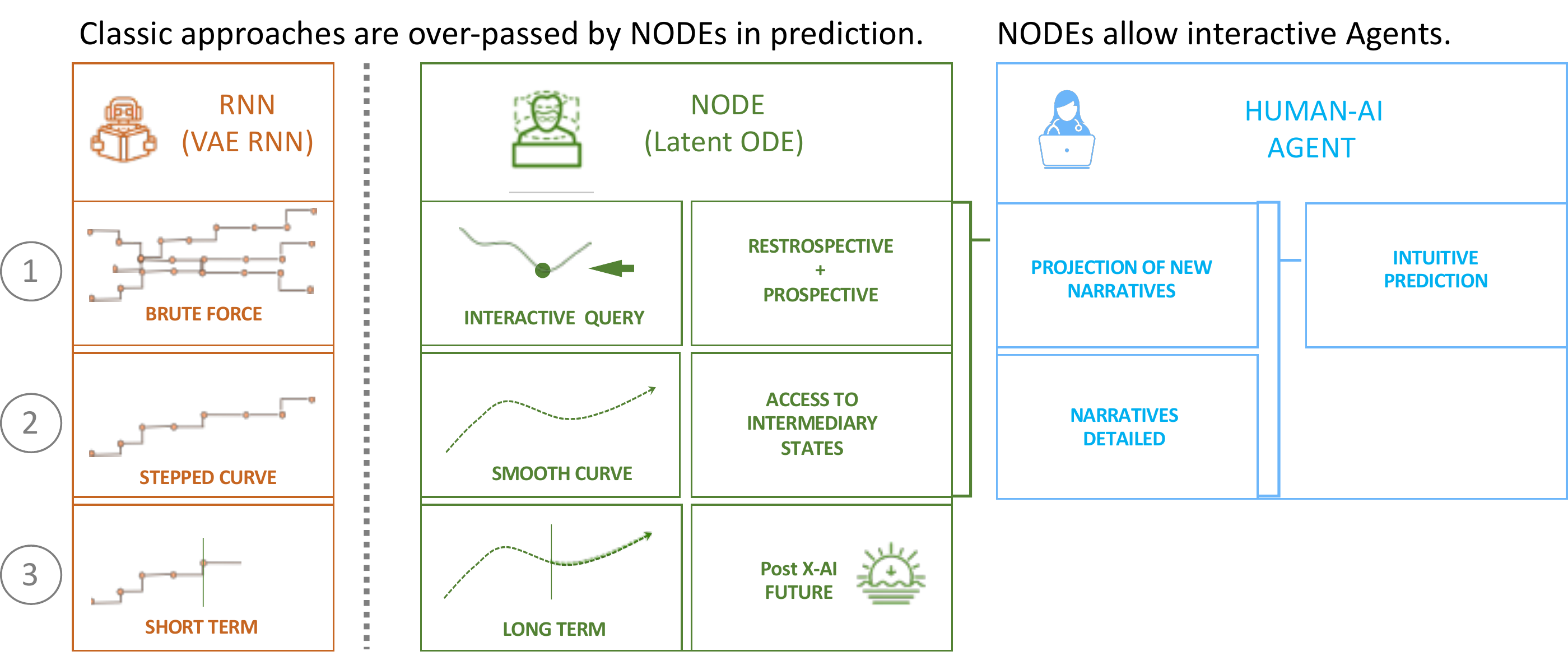}
    \caption{
    (1) RNNs deliver multiple future trajectories which require brute force analysis. NODEs offer interactive reconstruction of the past and future of the query point to intuit the plausibility of the new narrative.
    (2) RNNs make discrete predictions that do not allow the user to access intermediary states. NODEs help the understanding of intermediary states to rebuild a relevant narrative.
    (3) RNNs' horizon of predictability is short due to discrete predictions. NODEs give long term and highly accurate information without the need for explainability.
    An Interactive Agent will develop plausible narratives that support expert intuition to enhance the capacity to prevent disruptive changes.}
    \label{tab:shift}
\end{table*}

Hybrid Human-AI predictive systems could lead the way to a new generation of augmented decision-making solutions, and provide radical advances in readiness and response to still unpredictable events. Predictive agents built on intrinsically explainable ML architectures such as NODEs would offer objectivity when the rational foundations of a prediction are still disputed, and would provide dynamical representations to facilitate early adoption of humanly unpredictable scenarios, in the respect of the expert's world view. As an ultimate result, these proposed predictive agents could allow users to re-code nonrepresentational knowledge (i.e., intuition) into a dynamic representation of the data, thus leveraging the modeling power and advantage of differential equations.

A concrete application in the medical field could be the prediction of risks in Post Traumatic Stress Disorder (PTSD). PTSD is very difficult to model and project future states early, soon after a traumatic event (a time often referred to as the "blind zone"). Because many symptoms of PTSD are difficult to detect and measure (suffering, malaise, depression, suicidal thoughts, etc), creating models that make accurate prognoses is exponentially difficult. In our proposed system built on a NODE architecture, a predictive agent could encode the subject's evolution patterns into a NODE, and run a simulation of the possible future threats, providing then a concise description of the estimated risks. Thanks to a more accurate prediction, the physician, during the medical check-up, could decide faster whether to include or not the subject in a specific process of care.

\section*{Perspectives}

Achieving the vision of humans leveraging the predictive power of ML in a synergistic team relationship will take much planning and work, much of which is beyond mere model development. The first step should be to select and build models that are intrinsically understandable to human beings, and that naturally afford enhanced insight and support better decision making. We have demonstrated here one such system, built upon a robust and time-tested mathematical approach to modelling generative processes over time. Our demonstration, we hope, illustrates how the use of NODEs in medical prognoses is superior to any explanation attempt of black box models, and also supports user's natural intuition as a consequence of its design. 

\paragraph{Non-interpretable features:}

When the features are not intuitive to interpret for a given expert, it can be difficult to generate a narrative merely from extrapolated data. Doing so is the equivalent of attempting to convince someone of a different opinion or perspective- an effort with low historical likelihood of success. 
To help the construction of narratives and the interactions with a predictive agent, an interesting direction would be to extract additional variables of interest, that are distinct from the measured features. For instance, in the case of ICU patients, it could be interesting to have machine learning algorithms that extract from the latent trajectory the occurrence of specific events about different systems (respiratory, cardiac, etc.) categorized by physicians to help supporting narratives. The mechanistic representation of the expert decision making, even if incomplete, could contain, for example, mutually exclusive symptoms appearing in a time frame defined by physical bounds, i.e. critical event intensity.

An additional algorithm could be used to extract information from the intractable latent space to augment the basic information in expert knowledge. For this step, we could either use classical or powerful deep learning algorithms, since extracted data are not yet subject to explainability. Doing so could be framed as adding prior basic knowledge to the equation resolution. In the field of physics, to model systems conserving their total energy, it is possible to add an energy constraint to the NODE, to ensure that trajectories satisfy this condition. In technical terms, this is enforced using a Hamiltonian structure on the NODE, and the corresponding machine learning algorithm is studied in depth in \cite{Symoden}. This sensibility to prior knowledge needs to be investigated, in particular for real world datasets.

To confirm the usefulness of these additionally extracted variables , it would then be necessary to conduct trials: the recommendation system would be tested by experts with or without this add-on and evaluated for machine prediction acceptability. This is our proposed plan for the future.

\paragraph{}
In conclusion, we have demonstrated the potential utility of using NODE architecture on real-life data to enhance and improve human prognosis in medical decisions. We have illustrated the benefits, both intrinsic and designed, of such an architecture, and have discussed why these benefits are likely to enhance human-machine teaming and technology acceptance of ML in expert domains such as medicine.

\vspace{1.3cm}
    \textit{This work was supported by the US Office of Naval Research Global : ONRG  - Research Grant - [N62909-20-1-2076].}

\newpage

\printbibliography

\end{document}